\documentclass[10pt,twocolumn,letterpaper]{article}

\usepackage[pagenumbers]{cvpr} % To force page numbers, e.g. for an arXiv version

\usepackage[accsupp]{axessibility}

\definecolor{cvprblue}{rgb}{0.21,0.49,0.74}
\usepackage[pagebackref,breaklinks,colorlinks,allcolors=cvprblue]{hyperref}
\usepackage{hyperref}
\usepackage{url}
\usepackage{marvosym} % 提供 \Letter 小信封图标
\usepackage{amsmath}  % 提供 \text{} 环境，让小信封在角标中正常显示
\usepackage[utf8]{inputenc} % allow utf-8 input
\usepackage[T1]{fontenc}    % use 8-bit T1 fonts
\usepackage{hyperref}       % hyperlinks
\usepackage{url}            % simple URL typesetting
\usepackage{booktabs}       % professional-quality tables
\usepackage{amsfonts}       % blackboard math symbols
\usepackage{nicefrac}       % compact symbols for 1/2, etc.
\usepackage{microtype}      % microtypography
\usepackage{xcolor}         % colors
\usepackage{multirow}
\usepackage{caption}
\usepackage{subcaption}
\usepackage{amssymb}  % 方便使用 \checkmark
\usepackage{pifont}   % 方便使用 \ding{51}, \ding{55} 等符号
\usepackage{graphicx} % 如果需要 \resizebox
\usepackage{adjustbox}% 如果需要 \begin{adjustbox}{max width=\textwidth}
\usepackage{tikz}       % 用于绘制圆圈和其它形状
\usepackage{makecell}
\usepackage{marvosym}

\definecolor{GreenPartA}{HTML}{72CC58}
\definecolor{GreenPartB}{HTML}{4FC032}
\definecolor{RedPartA}{HTML}{ED6535}
\definecolor{RedPartB}{HTML}{E84108}

\newcommand{\cmark}{%
    \tikz[baseline=-0.6ex]{%
        \draw[fill=none, draw=none] (0,0) circle (0.95ex);

        \begin{scope}
        \clip (0,0) circle (0.95ex);
        \fill[GreenPartA] (0,0) -- (91:0.95ex) arc[start angle=91, end angle=-121, radius=0.95ex] -- cycle;
        \fill[GreenPartB] (0,0) -- (90:0.95ex) arc[start angle=90, end angle=240, radius=0.95ex] -- cycle;
        \end{scope}

        \draw[line cap=round, line join=round, white, line width=0.12em]
             (-0.45ex, 0.00ex) -- (-0.15ex, -0.30ex) -- (0.45ex, 0.35ex);
    }%
}

\newcommand{\xmark}{%
    \tikz[baseline=-0.6ex]{%
        \draw[fill=none, draw=none] (0,0) circle (0.95ex);

        \begin{scope}
        \clip (0,0) circle (0.95ex);
        \fill[RedPartA] (0,0) -- (91:0.95ex) arc[start angle=91, end angle=-121, radius=0.95ex] -- cycle;
        \fill[RedPartB] (0,0) -- (90:0.95ex) arc[start angle=90, end angle=240, radius=0.95ex] -- cycle;
        \end{scope}

        \draw[line cap=round, line join=round, white, line width=0.12em]
            (-0.4ex, -0.4ex) -- (0.4ex, 0.4ex);
        \draw[line cap=round, line join=round, white, line width=0.12em]
            (0.4ex, -0.4ex) -- (-0.4ex, 0.4ex);
    }%
}

\def\paperID{14715} % *** Enter the Paper ID here
\def\confName{CVPR}
\def\confYear{2026}

\title{CompBench: Benchmarking Complex Instruction-guided Image Editing}

\author{
    Bohan Jia$^{1,*}$, Wenxuan Huang$^{1,4,*}$, Yuntian Tang$^{1,*}$, Junbo Qiao$^{1}$, Jincheng Liao$^{1}$, \\
    Shaosheng Cao$^{2,\text{\Letter}}$, Fei Zhao$^{2}$, Zhaopeng Feng$^{5}$, Zhouhong Gu$^{6}$, Zhenfei Yin$^{7}$, \\
    Lei Bai$^{8}$, Wanli Ouyang$^{4}$, Lin Chen$^{9}$, Fei Zhao$^{10}$, Yao Hu$^{2}$, Zihan Wang$^{1}$, Yuan Xie$^{1}$, Shaohui Lin$^{1,3,\text{\Letter}}$ \\
    \vspace{0.15cm}
    $^1$East China Normal University, $^2$Xiaohongshu Inc., $^3$KLATASDS, MOE, China \\
    $^4$The Chinese University of Hong Kong, $^5$Zhejiang University, $^6$Fudan University, $^7$University of Oxford \\
    $^8$Shanghai Jiao Tong University, $^9$University of Science and Technology of China, $^{10}$Nanjing University \\
    \vspace{0.1cm}
    {\tt\small 51275901134@stu.ecnu.edu.cn} \\
    \vspace{0.1cm}
    \small $^*$Equal contribution \quad $^\text{\Letter}$Corresponding author \quad $^\dagger$Project leader
}

\begin{document}
\maketitle
\begin{abstract}
While real-world applications increasingly demand intricate scene manipulation, existing instruction-guided image editing benchmarks often oversimplify task complexity and lack comprehensive, fine-grained instructions. To bridge this gap, we introduce \textbf{\textit{CompBench}}, a large-scale benchmark specifically designed for \textit{complex instruction-guided image editing}. CompBench features challenging editing scenarios that incorporate fine-grained instruction following, spatial and contextual reasoning, thereby enabling comprehensive evaluation of image editing models' precise manipulation capabilities.
To construct CompBench, we propose an MLLM-human collaborative framework with tailored task pipelines. Furthermore, we propose an instruction decoupling strategy that disentangles editing intents into four key dimensions: location, appearance, dynamics, and objects, ensuring closer alignment between instructions and complex editing requirements. Extensive evaluations reveal that CompBench exposes fundamental limitations of current image editing models and provides critical insights for the development of next-generation instruction-guided image editing systems. Our project page is available at \url{https://comp-bench.github.io/}.
\end{abstract}
    
\section{Introduction}
\label{sec:intro}

\begin{figure*}[h]
\centering
\vskip -0.2in
\includegraphics[width=0.82\textwidth]{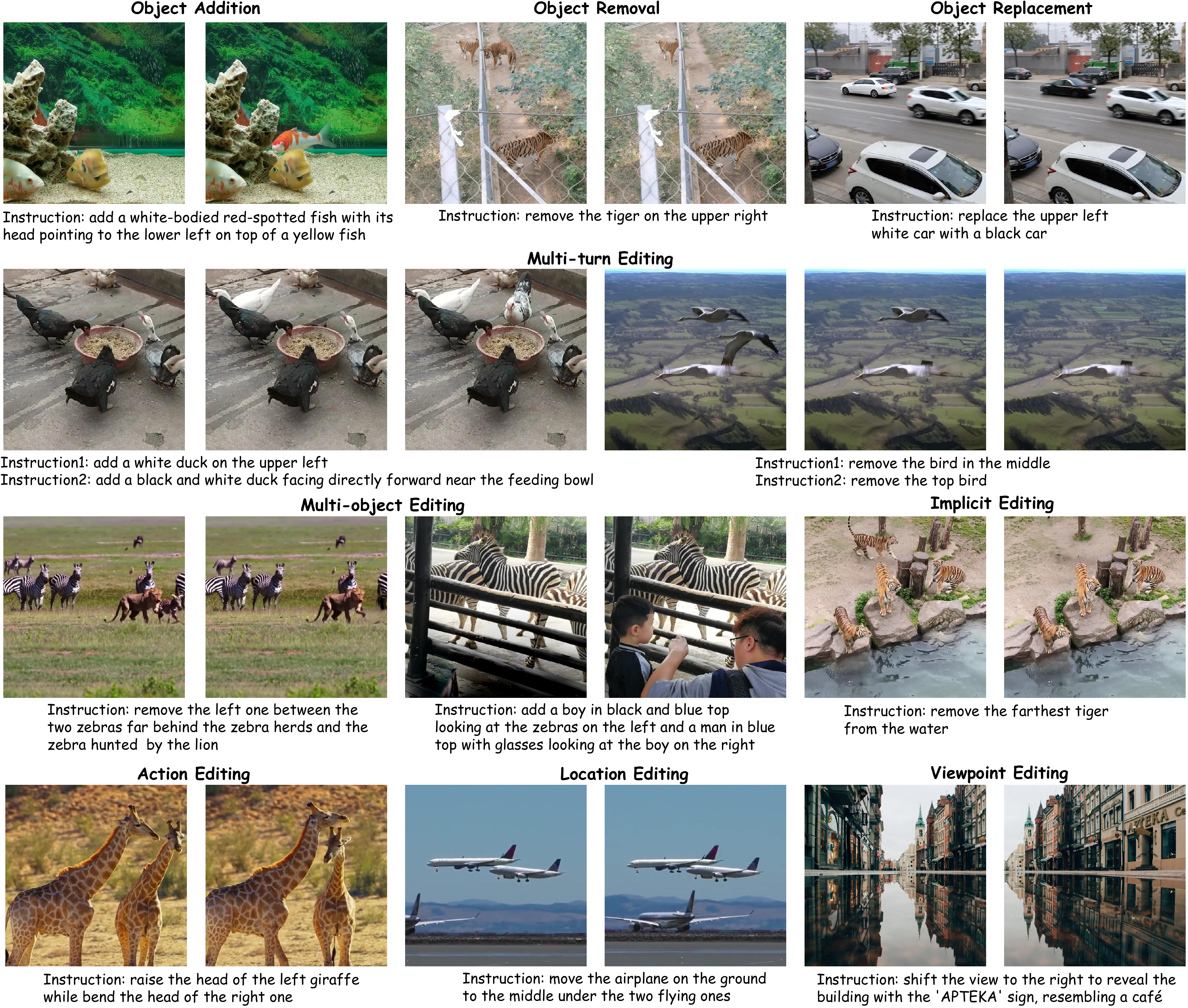}
\vskip -0.05in
\caption{\textbf{Examples of CompBench.}~The figure showcases nine tasks in our CompBench: object addition, object removal, object replacement, multi-object editing, multi-turn editing, implicit reasoning, action editing, location editing and viewpoint editing. }
\label{fig:compbench_expample}
\vskip -0.15in
\end{figure*}

\begin{figure*}[t]
\centering
% \vskip -0.2in
\includegraphics[width=0.99\textwidth]{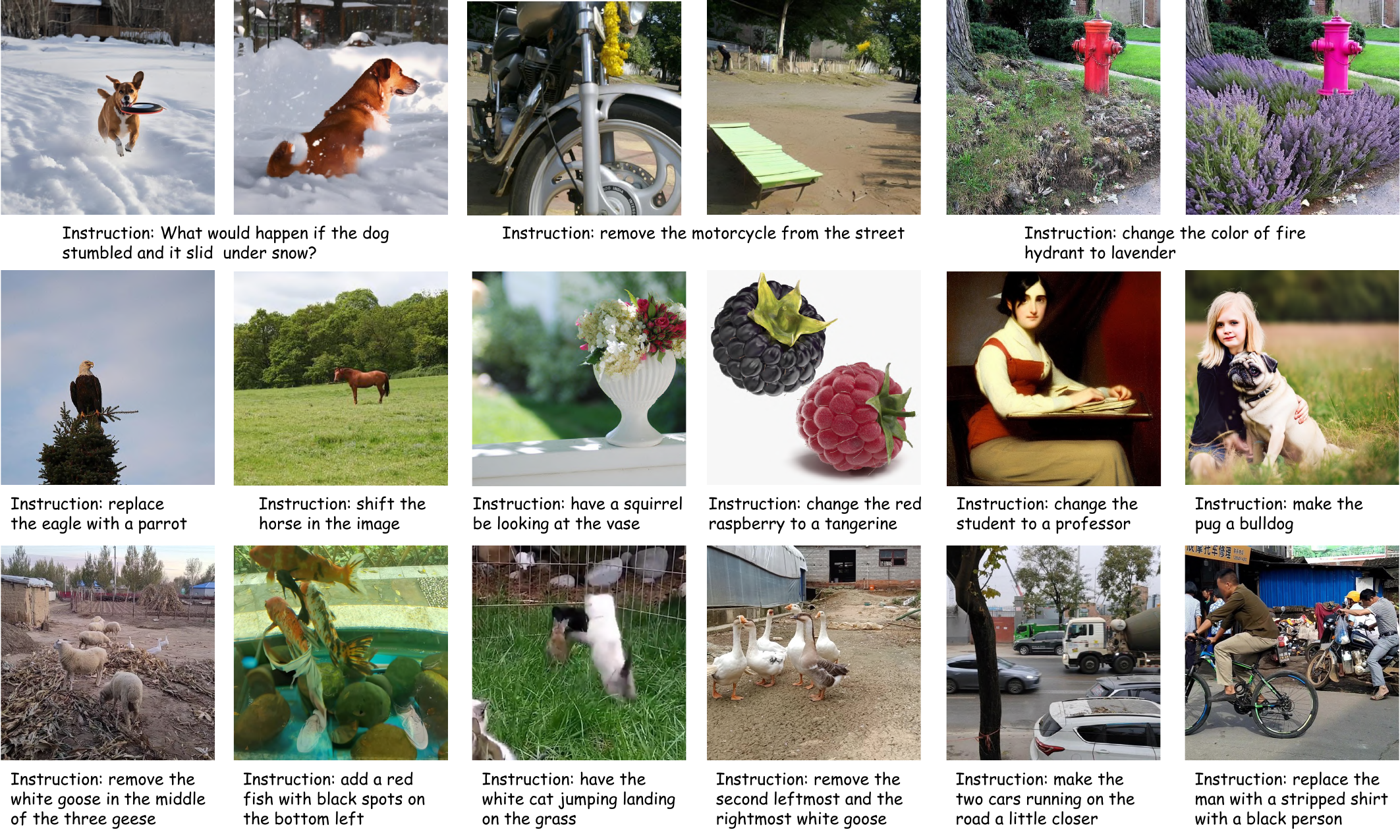}
% \vskip -0.05in
\caption{\textbf{Comparison between current datasets or benchmarks and our CompBench.} \textbf{First row:} failed cases of other benchmarks. These results fail to maintain background consistencies or introduce noticeable artifacts into the editing region.  \textbf{Second row:} Examples of other benchmarks. These cases lack scene complexity and instruction comprehensiveness. \textbf{Third row:} Examples of our CompBench. Our benchmark features complex real-world scenarios with precise instructions.}
\label{fig:bad_cases}
% \vskip -0.15in
\end{figure*}

Recent advances in instruction-guided image editing have pursued user-friendly and efficient manipulation of visual content. While such systems aim to simplify complex editing workflows, real-world applications often demand intricate instructions including spatial relationships, appearance details, and implicit reasoning.
This necessitates the development of models with comprehensive capabilities in visual grounding, contextual understanding, and complex reasoning, thereby presenting substantial challenges to existing methodologies.
However, as demonstrated in Figure~\ref{fig:bad_cases}, existing instruction-guided image editing benchmarks~\citep{sheynin2024emu, zhang2023magicbrush, huang2024smartedit} exhibit critical limitations in assessing these essential capabilities, primarily in three aspects:

% \textbf{Lack of Scene Complexity.}~A key limitation of current benchmarks is their insufficient scene complexity, which hampers the representation of intricate visual structures inherent in real-world images.
% This stems from two main factors.

% First, the prevalent use of synthetic images from text-to-image generation models, such as Stable Diffusion~\citep{stable-diffusion}, in previous benchmark construction~\citep{any-edit, i2ebench} results in scenes with sparse spatial layouts, limited foreground object diversity, minimal occlusions, and simplistic textures and lighting conditions. Such artificial compositions lack dense object interactions, natural clutter, and photorealistic qualities essential for evaluating practical editing capabilities. Even when incorporating real images from datasets, such as COCO~\citep{COCO}, these benchmarks often present oversimplified scenarios with elementary compositions insufficient for evaluating models on complex spatial relationships and interactions among multiple objects.

\paragraph{Lack of Scene Complexity.} A key limitation of current benchmarks is their insufficient scene complexity, which hampers the representation of intricate visual structures inherent in real-world images. Specifically, recent benchmark constructions~\citep{yu2025anyedit, ma2024i2ebench} predominantly source their images from general-purpose datasets, such as MS COCO~\citep{lin2014microsoft}. While these benchmarks utilize real images, they often present oversimplified, object-centric scenarios with elementary compositions. These scenes typically feature sparse spatial layouts, limited foreground object diversity, and minimal occlusions, lacking the dense object interactions and natural clutter essential for evaluating practical editing capabilities. Consequently, they remain insufficient for comprehensively evaluating models on complex spatial relationships and interactions among multiple objects.

This problem is further exacerbated by benchmark design choices, wherein
creators often deliberately exclude highly complex scenes featuring heavy occlusions, intricate details, or dynamic elements due to the challenges they pose for ground truth construction. 
While this practice facilitates more controllable evaluation, it creates a concerning discrepancy between benchmark performance and real-world applicability. 

Consequently, image editing models may attain high metric scores on these relatively simplified benchmarks, yet remain inadequate for real-world editing tasks that demand advanced scene understanding and manipulation.
For instance, in reasoning-based tasks, InstructPix2pix~\citep{brooks2023instructpix2pix} exhibits a notable performance decline on our CompBench compared with ReasonEdit~\citep{huang2024smartedit}, showing decreases of approximately 2.5 in PSNR, 0.02 in SSIM, and 0.4 in CLIP-Score.

\paragraph{Limited Instruction and Task Comprehensiveness.} Beyond their oversimplified visual scenes, current benchmarks are further constrained by the narrow scope of editing instructions and tasks, failing to reflect the complexity of real-world user demands. 
Most existing datasets rely on simplistic, atomic-level instructions (\textit{e.g.}, ``change the dog to a cat'') that lack contextual reasoning, and compositional logic typical of real user requests. In reality, user instructions often require complex reasoning and manipulation. These include multi-object editing (``remove the dog and the cat''), edits based on spatial relationships (``add a man to the right of the woman''), or action editing that modifies dynamic states (``make the man in white bend down more''). Current benchmarks, however, largely neglect these sophisticated task categories. This deficiency in instruction and task diversity prevents models from being rigorously tested on the full spectrum of challenges encountered in real-world applications. Consequently, their performance can be artificially inflated on simple tasks, providing an incomplete and misleading evaluation of true robustness and practical applicability.

\begin{table*}[t]
\centering
\caption{\textbf{Comparison of existing image-editing datasets and benchmarks.} 
Our benchmark supports seven core editing tasks, including multi-object, action and viewpoint editing, which are absent from most prior benchmarks.  
Scenario complexity is quantified by four indicators: \textit{Avg.\ Obj.} (average number of objects per image), \textit{Avg.\ Cat.} (average number of object categories per image), \textit{OCC} (percentage of images that contain occluded objects), and \textit{OOF} (percentage of images that contain out-of-frame objects). Across all four metrics, our benchmark exhibits the highest complexity, underscoring its suitability for rigorous evaluation.}
\label{tab:comparison}
\begin{adjustbox}{max width=\textwidth}
\begin{tabular}{lcc|ccccccc|cccc}
\hline
\multirow{2}{*}{\textbf{Datasets / Benchmarks}} &
\multirow{2}{*}{\textbf{Size}} &
\multirow{2}{*}{\textbf{Types}} &
\multicolumn{7}{c|}{\textbf{Task}} &
\multicolumn{4}{c}{\textbf{Complexity}} \\
\cline{4-14}
 &  &  &
\textbf{Local} & \textbf{Multi-turn} & \textbf{Multi-obj.} & \textbf{Implicit} & \textbf{Action} & \textbf{Location} & \textbf{Viewpoint} &
\textbf{Avg.\ Obj.} & \textbf{Avg.\ Cat.} & \textbf{Occ.\ Rate} & \textbf{OOF. Rate} \\
\hline
\multicolumn{14}{c}{\textit{Datasets}}\\
\hline
InstructPix2pix~\citep{brooks2023instructpix2pix}        & 313K  & 4  & \cmark & \xmark & \xmark & \xmark & \xmark & \xmark & \xmark & 8.71 & 4.16 & 79.36 & 81.39 \\
EditWorld~\citep{zeng2025editworld}                    & 8.6K  & 1  & \xmark & \xmark & \xmark & \cmark & \xmark & \xmark & \xmark & 8.01 & 4.45 & 76.67 & 72.00 \\
UltraEdit~\citep{zhao2024ultraedit}                    & 4M    & 9  & \cmark & \xmark & \xmark & \cmark & \xmark & \xmark & \xmark & 7.68 & 4.70 & 75.30 & 78.10 \\
SEED-Data-Edit~\citep{ge2024seed}          & 3.7M  & 6  & \cmark & \cmark & \cmark & \xmark & \xmark & \xmark & \xmark & 6.21 & 3.82 & 63.82 & 81.40 \\
HQ-Edit~\citep{hui2024hq}                        & 197K  & 6  & \cmark & \xmark & \xmark & \xmark & \xmark & \xmark & \xmark & 8.22 & 4.84 & 66.97 & 60.30 \\
AnyEdit~\citep{yu2025anyedit}                       & 2.5M  & 25 & \cmark & \xmark & \xmark & \cmark & \cmark & \cmark& \cmark & 6.95 & 4.37 & 60.45 & 57.20 \\
ImgEdit~\citep{ye2025imgedit}            & 1.2M  & 13  & \cmark & \xmark & \xmark & \xmark & \cmark & \xmark & \xmark & 9.01  & 4.72  & 69.65 & 69.14 \\
\hline
\multicolumn{14}{c}{\textit{Benchmarks}}\\
\hline
MagicBrush~\citep{zhang2023magicbrush}              & 10K   & 5  & \cmark & \cmark & \xmark & \xmark & \cmark & \xmark & \xmark & 9.22 & 5.04 & 91.71 & 78.30 \\
EMU\_Edit~\citep{sheynin2024emu}                   & --     & 8  & \cmark & \xmark & \xmark & \xmark & \xmark & \cmark & \xmark & 8.38 & 5.19 & 78.51 & 83.60 \\
Reason-Edit~\citep{huang2024smartedit}               & 0.2K  & -  & \cmark & \xmark & \xmark & \cmark & \xmark & \xmark & \xmark & 4.93 & 3.09 & 54.30 & 52.28 \\
I\textsuperscript{2}EBench~\citep{ma2024i2ebench}   & 2K    & 16 & \cmark & \xmark & \xmark & \xmark & \xmark & \xmark & \xmark & 7.03 & 4.20 & 68.78 & 66.40 \\
GEdit-Bench~\citep{liu2025step1x}               & 0.6K  & 11  & \cmark & \xmark & \xmark & \xmark & \cmark & \xmark & \xmark & 9.96  & 4.93 & 67.67  & 65.40 \\
Complex-Edit~\citep{yang2025complexedit}      & 1K  & 24  & \cmark & \xmark & \xmark & \cmark & \xmark & \xmark & \cmark & 9.23 & 4.77 & 78.29 & 72.98 \\
RefEdit~\citep{pathiraja2025refedit}   & 20K   & 5  & \cmark & \xmark & \cmark & \xmark & \xmark & \xmark & \xmark & 9.74 & 5.26 & 91.02 & 69.00 \\
KRIS-Bench~\citep{wu2025kris}  & 1.3K  & 22 & \cmark & \xmark & \cmark & \cmark & \xmark & \cmark & \cmark & 6.04 & 3.09 & 29.49 & 40.69 \\
ComplexBench-Edit~\citep{wang2025complexbench} & 763 & 10 & \cmark & \xmark & \cmark & \cmark & \cmark & \xmark & \xmark & 7.85 & 4.75 & 75.71 & 74.14 \\
\textbf{Ours}*                               & 3K    & 9  & \cmark & \cmark & \cmark & \cmark & \cmark & \cmark & \cmark & \textbf{13.58} & \textbf{5.87} & \textbf{98.47} & \textbf{86.38} \\
\hline
\end{tabular}
\end{adjustbox}
\end{table*}

\paragraph{Deficiencies in Edited Image Quality.} Another critical limitation of current benchmarks is the suboptimal quality of their edited images. Many existing datasets exhibit two predominant issues that compromise their reliability: (1) instruction-alignment inaccuracies, where the edited output fails to precisely fulfill the specified modifications. (2) conspicuous visual artifacts, such as geometric distortions, background inconsistencies, or semantically incoherent objects. These quality deficiencies introduce substantial noise into performance evaluations, potentially leading to misleading assessments of model capabilities. Consequently, such benchmarks may fail to effectively discriminate between truly sophisticated editing systems and those that merely produce superficially plausible but flawed results.

To address the aforementioned issues, we introduce \textbf{\textit{CompBench}}, the first large-scale benchmark for instruction-guided image editing in complex scenarios, specific examples are illustrated in Figure~\ref{fig:compbench_expample}. Our benchmark offers the following three major advantages:

\paragraph{Realistic and Complex Scene Composition.} As shown in Table~\ref{tab:comparison}, our benchmark encompasses scenes that embody the diverse and realistic complexities present in real-world settings. We compare CompBench with existing datasets and benchmarks across four dimensions: average number of objects, average number of object categories, overall object occlusion rate, and out-of-frame object rate. Details of these metrics are shown in the supplementary material. CompBench consistently surpasses prior benchmarks in all these metrics. Notably, our average number of objects per image is approximately \textbf{36.3\%} higher than the second best (GEdit-Bench~\citep{liu2025step1x}), demonstrating the heightened complexity and diversity of our scenes.

\paragraph{Comprehensive Task Coverage and High Difficulty Level.} As depicted in Figure~\ref{fig:overall tasks}, CompBench encompasses five major categories, consisting of local editing, multi-editing, action editing, scene spatial editing, and complex reasoning, spanning a total of nine tasks. These tasks are designed to challenge six core capabilities, with a detailed analysis of our benchmark's difficulty for each provided in the supplementary material. Additionally, we propose an Instruction Decomposition Strategy to improve the clarity and precision of image editing instructions. Specifically, we structure editing instructions along four dimensions: spatial positioning (\textit{e.g.}, ``left of the table''), visual attributes (such as color or texture), motion states (\textit{e.g.}, ``flying''), and object entities. This structured approach converts potentially ambiguous requests into well-defined specifications without sacrificing the natural expressiveness of instructions. By systematically covering each aspect of an editing operation while preserving the flexibility of natural language, our method produces instructions that are both intuitively understandable and technically precise for complex image editing tasks.

\paragraph{High-Quality Data Curation.}~Every sample in CompBench is meticulously constructed through multiple rounds of expert review, ensuring the highest quality of edits. Unlike other benchmarks where editing failures are common, all data in CompBench represent successfully executed editing results, with SSIM (Structural Similarity Index Measure) scores significantly outperforming those of other datasets, as illustrated in Figure~\ref{fig:bubble}. This rigorous quality control ensures that CompBench provides a reliable assessment of model performance in realistically complex editing scenarios.
\section{Related Works}
\paragraph{Instruction-guided Image Editing.} Instruction-guided image editing enables efficient image manipulation using only textual editing instructions, eliminating the need for manual mask or explicit visual inputs and better aligning with user intent. Diffusion models~\citep{ho2020denoising}, particularly Stable Diffusion~\citep{rombach2022high} (SD), facilitate this task significantly by supporting explicit text inputs. Methods built upon diffusion models such as InstructPix2pix~\citep{brooks2023instructpix2pix}, have greatly improved editing effectiveness. InstructPix2pix leverages large language models (LLMs)~\citep{vaswani2017attention, devlin2019bert, brown2020language, touvron2023llama} and text-to-image (T2I)~\citep{ramesh2021zero, ramesh2022hierarchical, rombach2022high, saharia2022photorealistic} models to generate large-scale datasets and trains a diffusion model that is capable of following natural language instructions. HIVE~\citep{zhang2024hive} introduces a reward model that leverages human feedback to align edits with human preferences. Approaches such as SmartEdit~\citep{huang2024smartedit}, MGIE~\citep{fu2023guiding}, and Step1X-Edit~\citep{liu2025step1x} integrate image and instruction representations using multi-modal large language models (MLLMs)~\citep{li2023blip, alayrac2022flamingo, liu2023visual, wang2024qwen2}, injecting these capabilities into diffusion models for more precise control. AnyEdit~\citep{yu2025anyedit} constructs an extremely large-scale multi-task dataset and adopts a mixture-of-experts (MoE)~\citep{fedus2022switch, du2022glam} architecture to better accommodate diverse editing tasks. SEED-X~\citep{ge2024seed} utilizes a visual tokenizer to unify image comprehension and generation, establishing a unified multi-granularity comprehension and generation model that enhances editing performance. GoT~\citep{fang2025got} incorporates Generation Chain-of-Thought~\citep{wei2022chain} reasoning into the editing process, allowing for more refined, step-by-step edits. Recently, FLUX.1 Kontext~\citep{labs2025flux} applies flow matching to build a unified image generation and editing model. Bagel~\citep{deng2025emerging} adopts a decoder only architecture to construct a multimodal understanding and generation model. Qwen-Image-Edit~\citep{wu2025qwen}, the editing model of Qwen-Image~\citep{wu2025qwen}, demonstrates strong text rendering and image editing capabilities.

\paragraph{Image Editing Benchmarks.} High-quality image editing datasets and benchmarks are crucial for model training and evaluation. Several notable benchmarks have been proposed: MagicBrush~\citep{zhang2023magicbrush} provides a manually curated 10K dataset covering single-turn, multi-turn, mask-provided, and mask-free editing tasks. EMU-edit~\citep{sheynin2024emu} introduces a challenging benchmark comprising seven diverse editing tasks. HQ-Edit~\citep{hui2024hq} employs a scalable data collection pipeline to create a high-quality dataset of 200K instruction-guided image editing samples. SmartEdit~\citep{huang2024smartedit} introduces Reason-Edit, a small-scale, manually curated benchmark focused on complex instruction-based image editing. Edit-world~\citep{zeng2025editworld} presents the concept of world-instructed image editing and creates a dataset featuring instructions in a world context. I2EBench~\citep{ma2024i2ebench} proposes a comprehensive evaluation benchmark with automated multi-dimensional assessment. UltraEdit~\citep{zhao2024ultraedit} develops a scalable framework for producing large and high-quality image editing datasets, introducing a large-scale instruction-based dataset. SEED-Data-Edit~\citep{ge2024seed} provides a hybrid dataset composed of auto-generated, real-world, and human-annotated multi-turn editing samples. More recently, ImgEdit~\citep{ye2025imgedit} introduces a large scale image editing dataset and a benchmark with multiple aspects. Step1X-Edit~\citep{liu2025step1x} construct GEdit-Bench~\citep{liu2025step1x} featuring real-world user instructions. Complex-Edit~\citep{yang2025complexedit} adopts a ``Chain-of-Edit'' pipeline to develop an image editing benchmark across instructions of different complexity. ComplexBench-Edit~\citep{wang2025complexbench} addresses combinatorial reasoning challenges by introducing a benchmark for chain-dependent instructions and a region-specific consistency metric. KRIS-Bench~\citep{wu2025kris} evaluates the cognitive capabilities of models through a knowledge-based taxonomy spanning factual, conceptual, and procedural dimensions. Furthermore, RefEdit~\citep{pathiraja2025refedit} targets the precise editing of specific objects in complex, multi-entity scenes based on referring expressions.

\begin{figure*}[t!]
\centering
\includegraphics[width=1.0\textwidth]{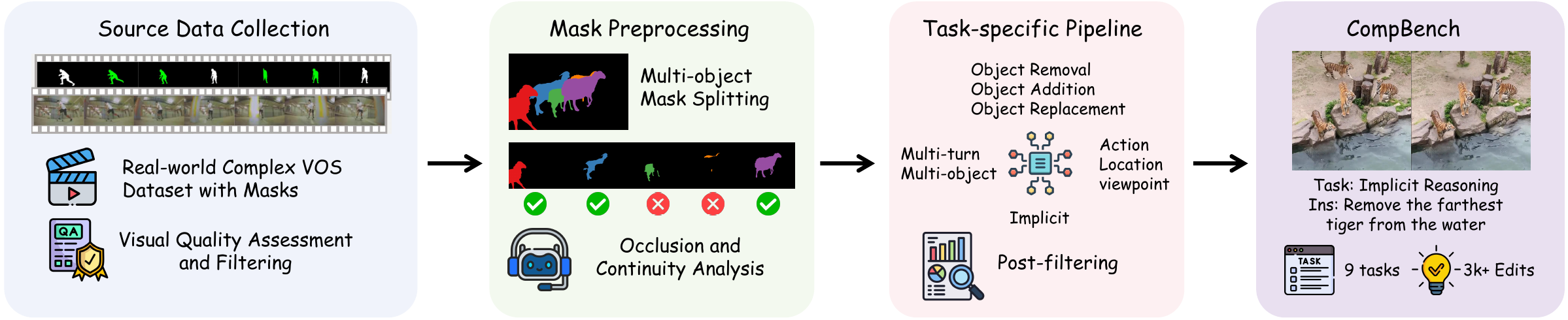}
\caption{\textbf{The construction pipeline of CompBench.} The pipeline consists of two main stages:
(a) Source data collection and preprocessing, wherein high-quality data are identified through image quality filtering, mask decomposition, occlusion and continuity evaluation, followed by thorough human verification. (b) Task-specific data generation using four specialized pipelines within our MLLM-Human Collaborative Framework, where multimodal large language models generate initial editing instructions that are subsequently validated by humans to ensure high-fidelity, semantically aligned instruction-image pairs for complex editing tasks.}
\label{fig:pipeline}
% \vskip -0.2in
\end{figure*}

\section{CompBench}
\subsection{Task Categorization and Definitions}
Our complex instruction-guided image editing benchmark, \textbf{CompBench}, contains over 3k image-instruction pairs. For comprehensive evaluation, we categorize the tasks into five major classes encompassing nine specific tasks: (1) Local Editing: manipulating objects via removal, addition, or replacement. (2) Multi-editing: addressing interactions across multi-turn or multi-object editing. (3) Action Editing: modifying dynamic states or object interactions. (4) Scene Spatial Editing: altering spatial properties through location or viewpoint editing. (5) Complex Reasoning: performing implicit contextual edits that require logical reasoning. Examples are illustrated in Figure~\ref{fig:compbench_expample}.

\subsection{Dataset Generation}
\label{sec:data_process}
In this section, we detailedly demonstrate the generation process of our CompBench(Figure~\ref{fig:pipeline}).

\paragraph{Source Data Collection and Preprocessing.} We utilize the MOSE dataset~\citep{ding2023mose} to address the lack of complex editing data. Image quality is first guaranteed by filtering corrupted frames via automated metrics (\textit{e.g.}, NIQE~\citep{zhang2015feature}) and subsequent manual verification. For masks, we decompose multi-object annotations into single-object instances. Finally, discontinuous or heavily occluded masks are discarded using MLLMs (\textit{e.g.}, Qwen-VL~\citep{wang2024qwen2}), followed by manual refinement to ensure pixel-level precision.

\paragraph{Task-specific Data Generation Pipelines.} We design four specialized pipelines to cover diverse complex instruction-guided image editing scenarios: (1)  local editing pipeline for object-level manipulations (object removal, object addition, object replacement). (2) action/scene spatial editing pipeline for modifying object dynamics or scene perspectives (action editing, location editing, viewpoint editing). (3) complex reasoning pipeline for implicit contextual edits requiring reasoning (implicit reasoning). (4) multi-editing pipeline for multi-object and multi-turn editing tasks. All pipelines adopt a unified MLLM-Human Collaborative Framework: multimodal large language models (MLLMs)~\citep{li2023blip, alayrac2022flamingo, liu2023visual, wang2024qwen2} generate initial task-specific instructions by analyzing visual scenes and editing goals, followed by human validation to ensure instruction-image alignment and image editing fidelity. Unsuccessful edits are iteratively re-generated or discarded, retaining only high-fidelity samples. Detailed procedures are provided in the supplementary material.

\begin{figure}[t]
  \centering
  \includegraphics[height=5.2cm,keepaspectratio]{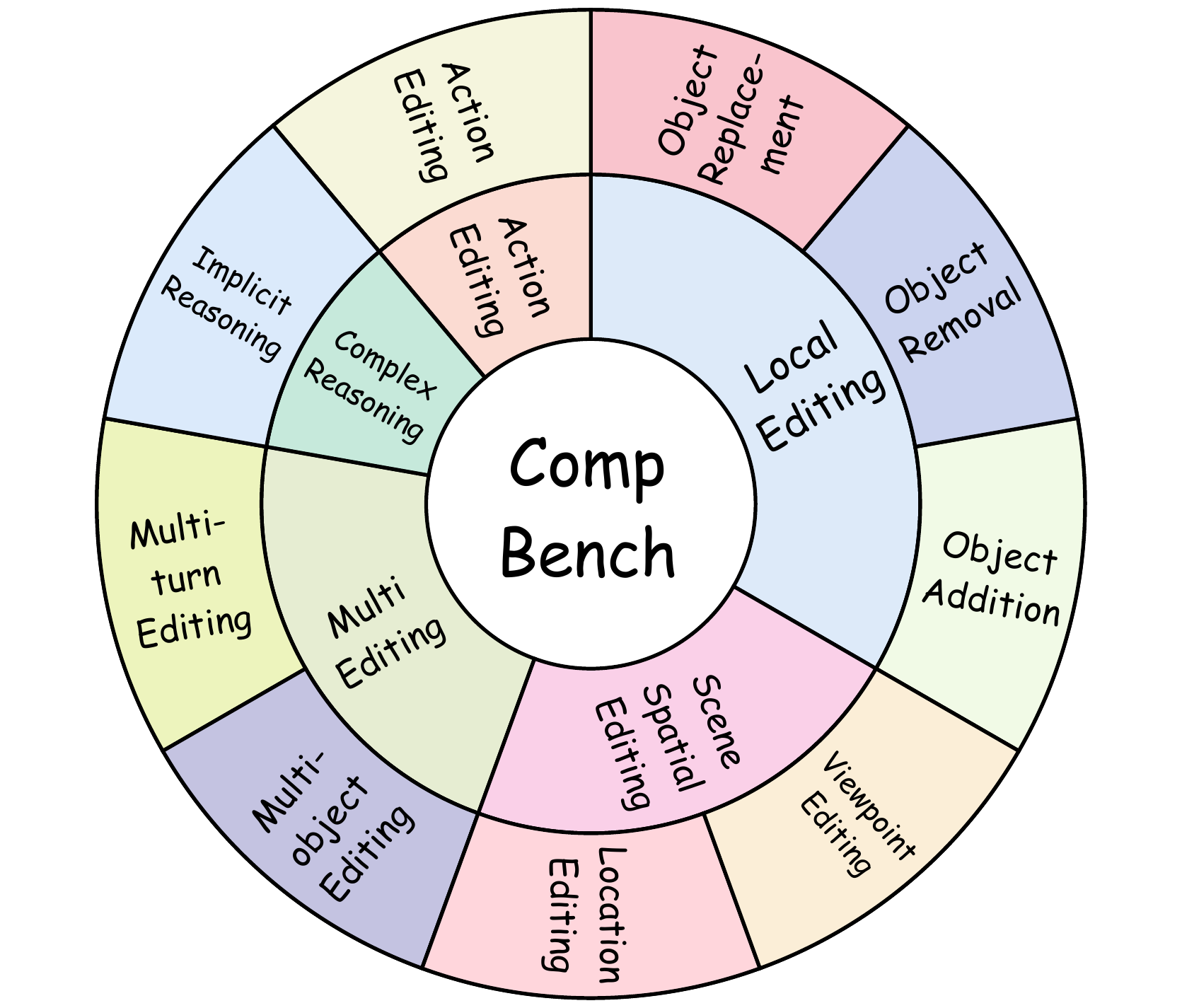}
  \caption{\textbf{Task taxonomy of CompBench.} Illustration of the full range of task types in CompBench.}
  \label{fig:overall tasks}
\end{figure}

\begin{figure}[t]
  \centering
  \includegraphics[height=5.2cm,keepaspectratio]{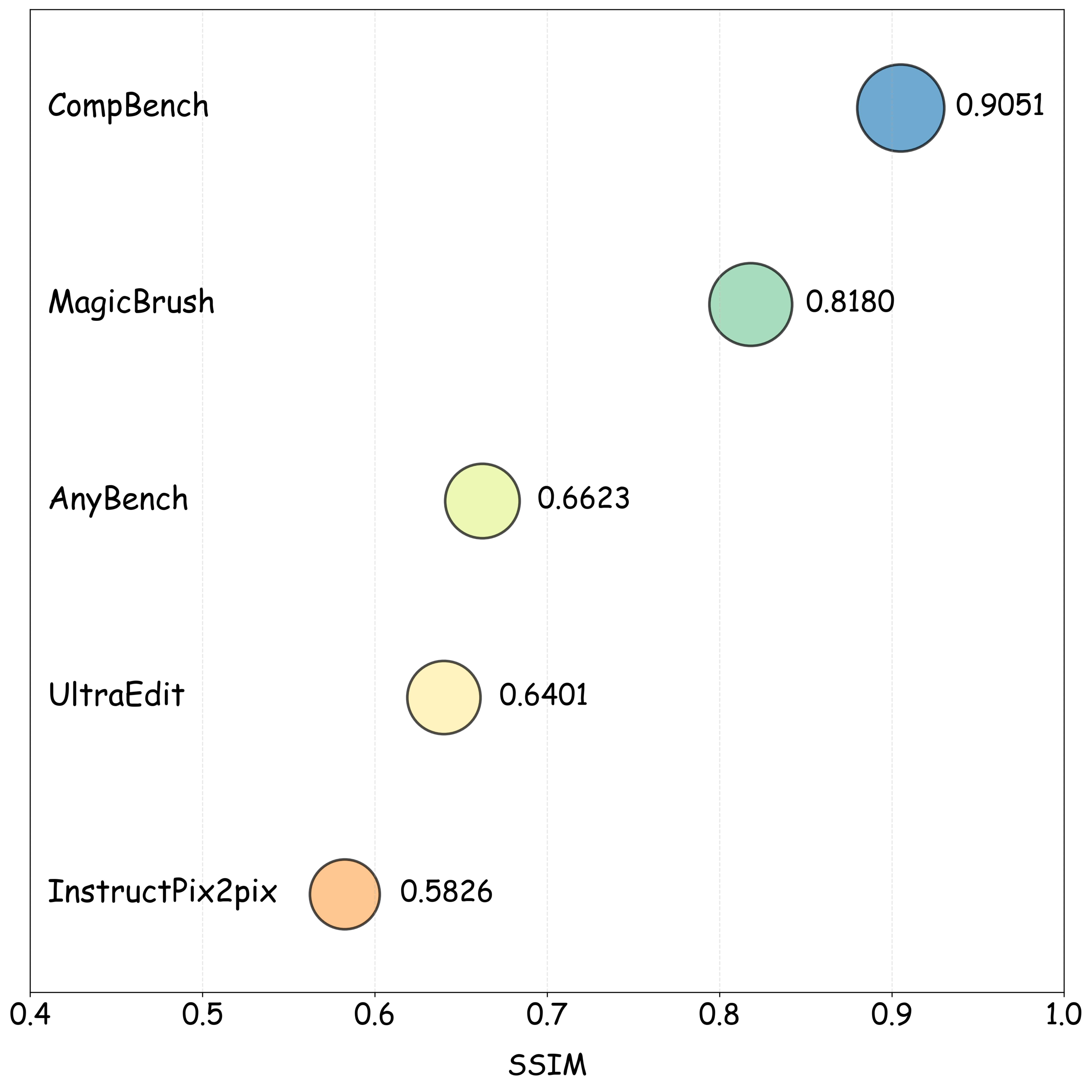}
  \caption{\textbf{SSIM comparison among different datasets and benchmarks.} Note that UltraEdit~\citep{zhao2024ultraedit} and InstructPix2pix~\citep{brooks2023instructpix2pix} are datasets, whereas the remaining entries are benchmarks.}
  \label{fig:bubble}
\end{figure}

\paragraph{Instruction Decomposition Strategy.} To enhance the clarity and precision of editing instructions, we propose a structured framework that organizes editing instructions along four aspects: spatial positioning, visual attributes, motion states, and object entities. This approach transforms ambiguous editing requests into well-defined specifications while maintaining natural expressiveness. The method employs a two-phase generation process: first, an MLLM produces dimension-aware instruction candidates by analyzing visual contexts. Then human experts refine these to ensure precision and consistency. By systematically addressing each aspect of the editing operation while preserving the flexibility of natural language, this framework enables the creation of instructions that are both intuitively understandable and technically precise for complex image editing tasks.

\paragraph{Characteristics and Statistics.} As illustrated in Figure~\ref{fig:overall tasks}, our benchmark comprises five major categories encompassing nine complex editing tasks, yielding over 3000 high-quality samples. We employ SSIM~\citep{wang2004image} to evaluate semantic consistency between pre- and post-edited images. As shown in Figure~\ref{fig:bubble}, CompBench achieves notably higher SSIM than other datasets and benchmarks. Notably, our dataset features significantly more challenging editing tasks requiring comprehensive capabilities such as visual grounding and complex reasoning. Quantitative indicators (\textit{e.g.}, average number of objects and categories) also demonstrate that our benchmark features substantially higher scene complexity. Detailed subtask definitions and core competency analyses are provided in the supplementary material.
\section{Experiments}
\label{sec:experiments}
\subsection{Settings}
\textbf{Baselines.}~Our study focuses on instruction-guided image editing models, excluding those based on global description guidance. The evaluated models include: InstructPix2pix~\citep{brooks2023instructpix2pix}, MagicBrush~\citep{zhang2023magicbrush}, HIVE~\citep{zhang2024hive}, Smart-edit~\citep{huang2024smartedit}, MGIE~\citep{fu2023guiding}, HQ-Edit~\citep{hui2024hq}, CosXL-Edit~\citep{cosxl}, UltraEdit~\citep{zhao2024ultraedit}, AnyEdit~\citep{yu2025anyedit}, Seed-X-Edit~\citep{ge2024seed}, GoT~\citep{fang2025got}, Step1X-Edit~\citep{liu2025step1x}, Bagel~\citep{deng2025emerging}, FLUX.1 Kontext~\citep{labs2025flux}, and Qwen-Image-Edit~\citep{wu2025qwen}.
% the recently released Qwen-Image-Edit~\citep{wu2025qwen}.

\begin{table*}[t]
\centering
\caption{\textbf{Evaluation results on local editing, multi-object editing and implicit reasoning.} LC-T denotes local CLIP scores between the edited foreground and the local description. LC-I refers to the CLIP image similarity between the foreground edited result and ground truth (GT) image. Top-three evaluation results are highlighted in \textcolor{red}{\textbf{red}} (1st), \textcolor{blue}{\textbf{blue}}(2nd), and \textcolor{green}{\textbf{green}} (3rd).}
\label{tab:all-editing}
\begin{adjustbox}{max width=1.0\textwidth}
\begin{tabular}{l|cc|ccc|cc|ccc|cc|ccc}
\hline
\multirow{3}{*}{\textbf{Model}} &
\multicolumn{5}{c|}{\textbf{Local Editing}} &
\multicolumn{5}{c|}{\textbf{Multi-object Editing}} &
\multicolumn{5}{c}{\textbf{Implicit Reasoning}}\\
\cline{2-16}
& \multicolumn{2}{c|}{\textbf{Foreground}} & \multicolumn{3}{c|}{\textbf{Background}} &
  \multicolumn{2}{c|}{\textbf{Foreground}} & \multicolumn{3}{c|}{\textbf{Background}} &
  \multicolumn{2}{c|}{\textbf{Foreground}} & \multicolumn{3}{c}{\textbf{Background}}\\
\cline{2-16}
& LC-T $\uparrow$ & LC-I $\uparrow$ & PSNR(dB) $\uparrow$ & SSIM $\uparrow$ & LPIPS $\downarrow$
& LC-T $\uparrow$ & LC-I $\uparrow$ & PSNR(dB) $\uparrow$ & SSIM $\uparrow$ & LPIPS $\downarrow$
& LC-T $\uparrow$ & LC-I $\uparrow$ & PSNR(dB) $\uparrow$ & SSIM $\uparrow$ & LPIPS $\downarrow$ \\
\hline
InstructPix2pix~\citep{brooks2023instructpix2pix} & 19.269 & 0.778 & 21.828 & 0.706 & 0.124 & 20.050 & 0.804 & 20.534 & 0.671 & 0.152 & 18.981 & 0.794 & 21.813 & 0.683 & 0.125  \\
MagicBrush~\citep{zhang2023magicbrush} & 20.051 & 0.798 & 23.429 & 0.741 & 0.090 & 20.004 & 0.821 & 24.176 & 0.738 & 0.081 & 19.498 & 0.826 & 22.143 & 0.714 & 0.106  \\
HIVE-w~\citep{zhang2024hive} & 19.804 & 0.771 & 19.903 & 0.641 & 0.198 & 19.949 & 0.782 & 19.896 & 0.624 & 0.210 & 18.590 & 0.777 & 20.261 & 0.602 & 0.219  \\
HIVE-c~\citep{zhang2024hive} & 19.226 & 0.773 & 21.732 & 0.689 & 0.147 & 19.904 & 0.798 & 21.854 & 0.676 & 0.144 & 18.888 & 0.787 & 22.167 & 0.666 & 0.144  \\
Smart-edit-7B~\citep{huang2024smartedit} & 19.999 & 0.799 & 24.389 & 0.761 & 0.074 & 20.186 & 0.825 & \textbf{\textcolor{green}{24.886}} & 0.745 & \textbf{\textcolor{green}{0.076}} & \textbf{\textcolor{blue}{19.740}} & 0.831 & 23.060 & 0.732 & 0.096  \\
MGIE~\citep{fu2023guiding} & 18.773 & 0.764 & 18.204 & 0.639 & 0.257 & 20.102 & 0.742 & 15.380 & 0.500 & 0.354 & 18.304 & 0.795 & 22.144 & 0.719 & 0.151  \\
CosXL-Edit~\citep{cosxl} & 19.068 & 0.778 & 20.809 & 0.712 & 0.148 & 19.606 & 0.807 & 21.068 & 0.698 & 0.153 & 18.003 & 0.800 & 21.190 & 0.683 & 0.156  \\
HQ-Edit~\citep{hui2024hq} & 18.888 & 0.734 & 11.768 & 0.411 & 0.428 & 19.401 & 0.708 & 13.032 & 0.432 & 0.400 & 18.688 & 0.766 & 11.877 & 0.387 & 0.459  \\
UltraEdit~\citep{zhao2024ultraedit} & 19.690 & 0.787 & 22.932 & 0.741 & 0.143 & 20.081 & 0.810 & 22.725 & 0.732 & 0.152 & 18.289 & 0.786 & 23.362 & 0.717 & 0.142  \\
AnyEdit~\citep{yu2025anyedit} & 19.994 & 0.797 & 22.812 & 0.716 & 0.124 & 20.257 & 0.810 & 23.493 & 0.710 & 0.116 & 19.572 & 0.816 & 20.276 & 0.639 & 0.191  \\
SEED-X~\citep{ge2024seed} & 17.900 & 0.780 & 21.456 & 0.805 & 0.139 & 19.418 & 0.835 & 21.166 & 0.798 & 0.148 & 17.437 & 0.782 & 21.438 & 0.790 & 0.134  \\
GoT~\citep{fang2025got} & 20.268 & 0.807 & 24.675 & 0.889 & \textbf{\textcolor{green}{0.067}} & 20.225 & 0.827 & 22.486 & 0.842 & 0.108 & 19.246 & 0.821 & \textbf{\textcolor{green}{24.889}} & 0.860 & 0.088  \\
Step1X-Edit~\citep{liu2025step1x} & 20.495 & 0.817 & 23.372 & 0.882 & 0.078 & 20.459 & 0.861 & 23.782 & \textbf{\textcolor{green}{0.886}} & 0.078 & 19.175 & 0.847 & 23.408 & \textbf{\textcolor{green}{0.869}} & \textbf{\textcolor{green}{0.083}}  \\
Bagel~\citep{deng2025emerging} & \textbf{\textcolor{green}{21.059}} & \textbf{\textcolor{red}{0.838}} & \textbf{\textcolor{red}{27.692}} & \textbf{\textcolor{blue}{0.935}} & \textbf{\textcolor{red}{0.045}} & \textbf{\textcolor{blue}{20.856}} & \textbf{\textcolor{red}{0.883}} & \textbf{\textcolor{red}{26.849}} & \textbf{\textcolor{blue}{0.932}} & \textbf{\textcolor{blue}{0.051}} & \textbf{\textcolor{green}{19.719}} & \textbf{\textcolor{red}{0.874}} & \textbf{\textcolor{red}{28.891}} & \textbf{\textcolor{blue}{0.919}} & \textbf{\textcolor{red}{0.052}}  \\
FLUX.1 Kontext~\citep{labs2025flux} & \textbf{\textcolor{blue}{21.328}} & \textbf{\textcolor{green}{0.821}} & \textbf{\textcolor{blue}{25.613}} & \textbf{\textcolor{red}{0.941}} & \textbf{\textcolor{blue}{0.050}} & \textbf{\textcolor{red}{21.020}} & \textbf{\textcolor{blue}{0.868}} & \textbf{\textcolor{blue}{26.278}} & \textbf{\textcolor{red}{0.952}} & \textbf{\textcolor{red}{0.048}} & 19.596 & \textbf{\textcolor{blue}{0.867}} & \textbf{\textcolor{blue}{25.401}} & \textbf{\textcolor{red}{0.932}} & \textbf{\textcolor{blue}{0.061}}  \\
Qwen-Image-Edit~\citep{wu2025qwen} & \textbf{\textcolor{red}{21.522}} & \textbf{\textcolor{blue}{0.828}} & \textbf{\textcolor{green}{24.968}} & \textbf{\textcolor{green}{0.892}} & 0.072 & \textbf{\textcolor{green}{20.722}} & \textbf{\textcolor{green}{0.864}} & 23.492 & 0.826 & 0.103 & \textbf{\textcolor{red}{20.046}} & \textbf{\textcolor{green}{0.859}} & 22.815 & 0.775 & 0.124  \\
\hline
\end{tabular}
\end{adjustbox}
\end{table*}

\begin{table*}[t]
\centering
\caption{\textbf{Evaluation results on multi-turn editing.}}
\label{tab:multi-turn-new}
\begin{adjustbox}{max width=0.75\textwidth}
\begin{tabular}{l|cc|ccc|cc|ccc}
\hline
\multirow{2}{*}{\textbf{Model}} &
\multicolumn{5}{c|}{\textbf{Turn1}} &
\multicolumn{5}{c}{\textbf{Turn2}} \\
\cline{2-11}
& \multicolumn{2}{c|}{Foreground} & \multicolumn{3}{c|}{Background}
& \multicolumn{2}{c|}{Foreground} & \multicolumn{3}{c}{Background} \\
\hline
 &\multicolumn{1}{c}{LC-T} & LC-I & PSNR & SSIM & LPIPS & \multicolumn{1}{c}{LC-T} & LC-I & PSNR & SSIM & LPIPS \\
\hline
InstructPix2pix~\citep{brooks2023instructpix2pix} & 19.617 & 0.786 & 21.361 & 0.682 & 0.134 & 19.856 & 0.777 & 17.764 & 0.573 & 0.233  \\
MagicBrush~\citep{zhang2023magicbrush} & 19.564 & 0.811 & 24.099 & 0.731 & 0.089 & 19.694 & 0.809 & 21.294 & 0.683 & 0.133  \\
HIVE-w~\citep{zhang2024hive} & 20.012 & 0.781 & 20.065 & 0.622 & 0.195 & \textbf{\textcolor{green}{20.389}} & 0.773 & 17.315 & 0.533 & 0.270  \\
HIVE-c~\citep{zhang2024hive} & 19.922 & 0.789 & 21.331 & 0.660 & 0.154 & 19.964 & 0.779 & 18.350 & 0.590 & 0.217  \\
Smart-edit-7B~\citep{huang2024smartedit} & 19.595 & 0.813 & 24.653 & 0.740 & 0.080 & 19.571 & 0.808 & \textbf{\textcolor{green}{23.383}} & 0.723 & \textbf{\textcolor{green}{0.104}}  \\
MGIE~\citep{fu2023guiding} & 19.194 & 0.814 & 18.473 & 0.602 & 0.246 & 19.328 & 0.803 & 14.820 & 0.493 & 0.366  \\
HQ-Edit~\citep{hui2024hq} & 19.364 & 0.758 & 12.384 & 0.391 & 0.410 & 19.228 & 0.753 & 11.608 & 0.351 & 0.478  \\
CosXL-Edit~\citep{cosxl} & 19.670 & 0.790 & 20.292 & 0.681 & 0.167 & 19.430 & 0.769 & 16.390 & 0.566 & 0.312  \\
UltraEdit~\citep{zhao2024ultraedit} & 19.816 & 0.793 & 23.065 & 0.737 & 0.151 & 20.015 & 0.793 & 19.927 & 0.666 & 0.208  \\
AnyEdit~\citep{yu2025anyedit} & 19.858 & 0.812 & 23.398 & 0.711 & 0.113 & 20.061 & 0.806 & 20.001 & 0.633 & 0.189  \\
SEED-X~\citep{ge2024seed} & 19.576 & 0.796 & 21.049 & 0.792 & 0.153 & 19.188 & 0.769 & 17.708 & 0.629 & 0.280  \\
GoT~\citep{fang2025got} & 19.833 & 0.815 & 24.840 & \textbf{\textcolor{green}{0.891}} & \textbf{\textcolor{green}{0.067}} & 19.832 & 0.809 & \textbf{\textcolor{blue}{23.487}} & \textbf{\textcolor{green}{0.855}} & 0.107  \\
Step1X-Edit~\citep{liu2025step1x} & 19.873 & 0.833 & 23.802 & 0.888 & 0.078 & 19.754 & 0.834 & 21.016 & 0.832 & 0.125  \\
Bagel~\citep{deng2025emerging} & \textbf{\textcolor{green}{20.060}} & \textbf{\textcolor{red}{0.852}} & \textbf{\textcolor{red}{28.959}} & \textbf{\textcolor{blue}{0.949}} & \textbf{\textcolor{red}{0.036}} & 20.007 & \textbf{\textcolor{red}{0.850}} & \textbf{\textcolor{red}{23.941}} & \textbf{\textcolor{blue}{0.897}} & \textbf{\textcolor{red}{0.085}}  \\
FLUX.1 Kontext~\citep{labs2025flux} & \textbf{\textcolor{red}{20.634}} & \textbf{\textcolor{blue}{0.845}} & \textbf{\textcolor{blue}{26.193}} & \textbf{\textcolor{red}{0.954}} & \textbf{\textcolor{blue}{0.046}} & \textbf{\textcolor{red}{20.620}} & \textbf{\textcolor{blue}{0.840}} & 22.417 & \textbf{\textcolor{red}{0.907}} & \textbf{\textcolor{blue}{0.094}}   \\
Qwen-Image-Edit~\citep{wu2025qwen} & \textbf{\textcolor{blue}{20.468}} & \textbf{\textcolor{blue}{0.845}} & \textbf{\textcolor{green}{24.863}} & 0.846 & 0.088  & \textbf{\textcolor{blue}{20.584}} & \textbf{\textcolor{green}{0.838}} & 20.814 & 0.783 & 0.151  \\
\hline
\end{tabular}
\end{adjustbox}
\end{table*}

\paragraph{Evaluation Metrics and Methods.} For local editing, multi-editing, and implicit reasoning tasks, we employ a foreground-background decoupling strategy. To evaluate background preservation, we compute PSNR, SSIM~\citep{wang2004image}, and LPIPS~\citep{zhang2018unreasonable} on the background regions. For foreground evaluation, we assess two aspects using CLIP~\citep{radford2021learning}: \textit{editing accuracy} via CLIP image similarity between the localized edited foreground and the ground truth (GT) image (denoted as LC-I), and \textit{instruction following} via local CLIP scores between the edited foreground and the local description (denoted as LC-T).

Additionally, for action, location, and viewpoint editing tasks—where the object's morphology, position, or viewpoint may change significantly—automatic metrics alone are insufficient. To address this, we introduce multi-perspective scoring via GPT-4o~\citep{gpt4o}, Qwen2.5-VL-72B~\citep{bai2025qwen25vltechnicalreport}, and human annotators. We design tailored prompts instructing the models to rate editing performance on a 0-10 scale. Simultaneously, trained human annotators evaluate background fidelity, editing intent, instruction following, and artifact presence based on standardized guidelines. Detailed prompts, annotation instructions, and additional evaluation results are available in the supplementary material.

\begin{table*}[t]
    \centering
    \caption{\textbf{Comparison on Action, Location, and Viewpoint Editing.} Results for GPT-4o, Qwen-72B, Human Evaluation, and Average scores (top-3 per column highlighted in red, blue, green).}
    \label{tab:gpt}
    \begin{adjustbox}{max width=0.8\textwidth}
    \begin{tabular}{l|rrrr|rrrr|rrrr}
    \hline
    \multirow{2}{*}{\textbf{Model}} &
    \multicolumn{4}{c|}{\textbf{Action}} &
    \multicolumn{4}{c|}{\textbf{Location}} &
    \multicolumn{4}{c}{\textbf{Viewpoint}} \\
    \cline{2-13}
    & GPT & Qwen & Human & Avg. & GPT & Qwen & Human & Avg. & GPT & Qwen & Human & Avg. \\
    \hline
    InstructPix2pix~\citep{brooks2023instructpix2pix} & 3.047 & 1.124 & 3.101 & 2.424 & 3.425 & 2.167 & 2.859 & 2.817 & 
    0.699 & 0.482 & 0.036 & 0.406 \\
    MagicBrush~\citep{zhang2023magicbrush} & 3.511 & 1.449 & 3.584 & 2.848 & 4.603 & 2.260 & 3.717 & 3.527 & 
    0.892 & 0.410 & 0.108 & 0.470 \\
    HIVE-w~\citep{zhang2024hive} & 3.151 & 1.764 & 3.067 & 2.661 & 4.110 & 2.192 & 3.421 & 3.241 & 
    1.494 & 0.283 & 0.036 & 0.604 \\
    HIVE-c~\citep{zhang2024hive} & 3.977 & 1.596 & 3.797 & 3.123 & 4.192 & 2.470 & 3.558 & 3.407 & 2.193 & 0.675 & 0.145 & 1.004 \\
    Smart-edit-7B~\citep{huang2024smartedit} 
      & 4.233 & 1.607 & 4.348 & 3.396
      & 3.890 & 2.875 & 3.505 & 3.423
      & 2.169 & 0.590 & 0.410 & 1.056 \\
    MGIE~\citep{fu2023guiding} & 1.921 & 1.213 & 1.797 & 1.644 & 1.726 & 1.795 & 1.728 & 1.750 
    & 0.205 & 0.193 & 0 & 0.133 \\
    CosXL-Edit~\citep{cosxl} 
      & 4.270 & 2.375 & 3.966 & 3.537
      & \textbf{\textcolor{green}{5.479}} & 2.493 & 4.517 & 4.163
      & 1.916 & 0.988 & 0.301 & 1.068 \\
    HQ-Edit~\citep{hui2024hq} & 1.449 & 0.528 & 1.033 & 1.003 & 1.425 & 0.726 & 1.079 & 1.077 & 0.470 & 0.289 & 0 & 0.253 \\
    UltraEdit~\citep{zhao2024ultraedit} & 4.449 & 1.807 & 4.235 & 3.497 & 4.014 & 2.055 & 3.410 & 3.160 & 0.494 & 0.706 & 0 & 0.400 \\
    AnyEdit~\citep{yu2025anyedit} & 3.750 & 0.978 & 3.168 & 2.632 & 5.068 & 2.479 & 4.178 & 3.908 & 1.687 & 0.783 & 0.072 & 0.847 \\
    SEED-X~\citep{ge2024seed} & 2.270 & 1.494 & 1.685 & 1.816 
    & 3.028 & 3.247 & 2.771 & 3.015 
    & 2.241 & 1.169 & 0 & 1.137 \\
    GoT~\citep{fang2025got} 
      & 3.337 & 1.989 & 3.134 & 2.820
      & 3.625 & 3.192 & 3.164 & 3.327
      & 0.916 & 0.675 & 0.446 & 0.679 \\
    Step1X-Edit~\citep{liu2025step1x}
      & \textbf{\textcolor{green}{6.270}} & \textbf{\textcolor{green}{3.944}} & \textbf{\textcolor{green}{5.348}} & \textbf{\textcolor{green}{5.187}}
      & 5.041 & \textbf{\textcolor{green}{4.479}} & \textbf{\textcolor{blue}{4.786}} & \textbf{\textcolor{green}{4.769}}
      & 2.470 & 1.205 & 0.663 & 1.446 \\
    Bagel~\citep{deng2025emerging}
      & \textbf{\textcolor{blue}{6.899}} & \textbf{\textcolor{blue}{5.056}} & \textbf{\textcolor{blue}{6.629}}  & \textbf{\textcolor{blue}{6.195}}
      & \textbf{\textcolor{red}{7.137}} & \textbf{\textcolor{red}{6.233}} & \textbf{\textcolor{red}{6.219}} & \textbf{\textcolor{red}{6.530}} 
      & \textbf{\textcolor{blue}{5.193}} & \textbf{\textcolor{blue}{3.892}} & \textbf{\textcolor{blue}{4.663}} & \textbf{\textcolor{blue}{4.583}} \\
    FLUX.1 Kontext~\citep{labs2025flux}
    & 5.169 & 3.202 & 4.517 & 4.296
    & 3.000 & 3.110 & 3.836 & 3.315
    & \textbf{\textcolor{green}{3.471}} & \textbf{\textcolor{green}{2.373}} & \textbf{\textcolor{green}{3.108}} & \textbf{\textcolor{green}{2.984}}   \\
    Qwen-Image-Edit~\citep{wu2025qwen}
    & \textbf{\textcolor{red}{6.910}} & \textbf{\textcolor{red}{5.382}} & \textbf{\textcolor{red}{6.764}} & \textbf{\textcolor{red}{6.352}}
    & \textbf{\textcolor{blue}{7.055}} & \textbf{\textcolor{blue}{5.096}} & \textbf{\textcolor{green}{4.658}} &  \textbf{\textcolor{blue}{5.603}}
    & \textbf{\textcolor{red}{6.193}} & \textbf{\textcolor{red}{4.470}} & \textbf{\textcolor{red}{6.181}} &  \textbf{\textcolor{red}{5.615}}  \\
    \hline
    \end{tabular}
    \end{adjustbox}
    \end{table*}

\subsection{Experiment Results}
The experimental results for local editing, multi-turn editing, multi-object editing, implicit reasoning, and action/location/viewpoint editing are presented in Tables~\ref{tab:all-editing}, ~\ref{tab:multi-turn-new}, and ~\ref{tab:gpt}, respectively. Our key analysis of the results are as follows:
(1) No model dominates across all tasks. Among all evaluated models, Bagel~\citep{deng2025emerging} emerges as the most prominent one, achieving top results in 18 out of 37 metrics (nearly 50\%) across 9 tasks. Notably, Bagel~\citep{deng2025emerging}, Qwen-Image-Edit~\citep{wu2025qwen}, and FLUX.1 Kontext~\citep{labs2025flux} consistently deliver superior performance, securing top-three rankings in the majority of metrics across most tasks, followed by Step1X-Edit~\citep{liu2025step1x}. In contrast, HQ-Edit~\citep{hui2024hq} demonstrates substantially inferior results in nearly all tasks. 
(2) For multi-turn editing tasks, all models exhibit a notable decline in background consistency metrics during the second editing round. Among them, SmartEdit~\citep{huang2024smartedit} maintains relatively robust performance in the second editing turn. 
(3) Bagel~\citep{deng2025emerging} consistently leads in the LC-I metric, reflecting its superior foreground fidelity. Additionally, it ranks high in background consistency, effectively preserving spatial and contextual information during editing.
(4) For the more challenging action/location/viewpoint editing tasks, Qwen-Image-Edit~\citep{wu2025qwen} and Bagel~\citep{deng2025emerging} perform comparably and significantly outperform most other models. Step1X-Edit~\citep{liu2025step1x} also exhibits promising editing performance in these scenarios.
% \section{Insights}
% In this section, we investigate the underlying factors that lead to varying performances among different models on our proposed CompBench, and offer perspectives on future research directions for the field of image editing.

\begin{figure*}[htbp]
  \centering
    \begin{subfigure}{0.45\linewidth}
        \includegraphics[height=5.2cm,keepaspectratio]{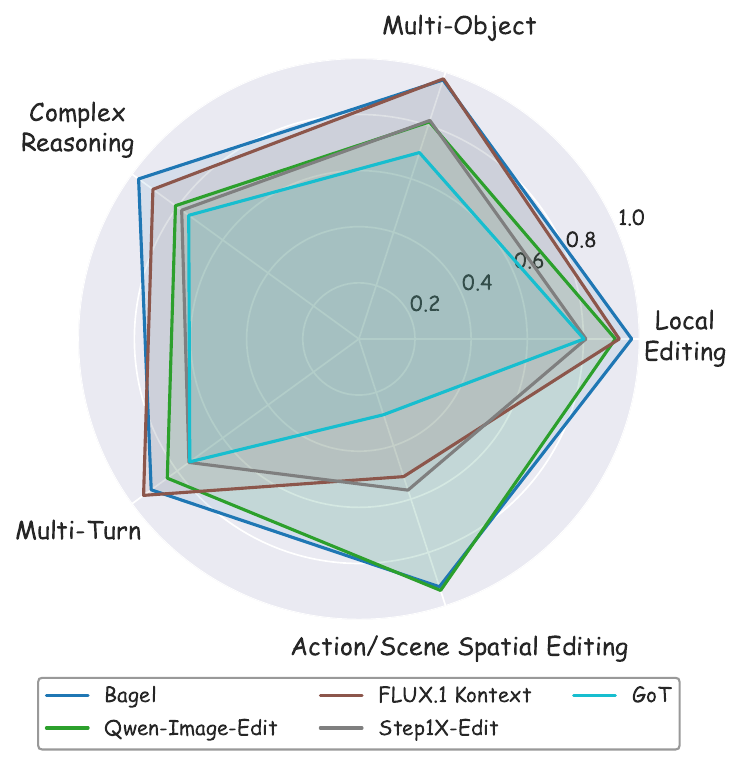}
        \caption{}
        \label{fig:radar}
    \end{subfigure}
 \hspace{0.00\linewidth} % 控制两图之间距离，可调小靠近
  %-- 子图 (a) --
  \begin{subfigure}{0.45\linewidth}
    \includegraphics[height=5.2cm,keepaspectratio]{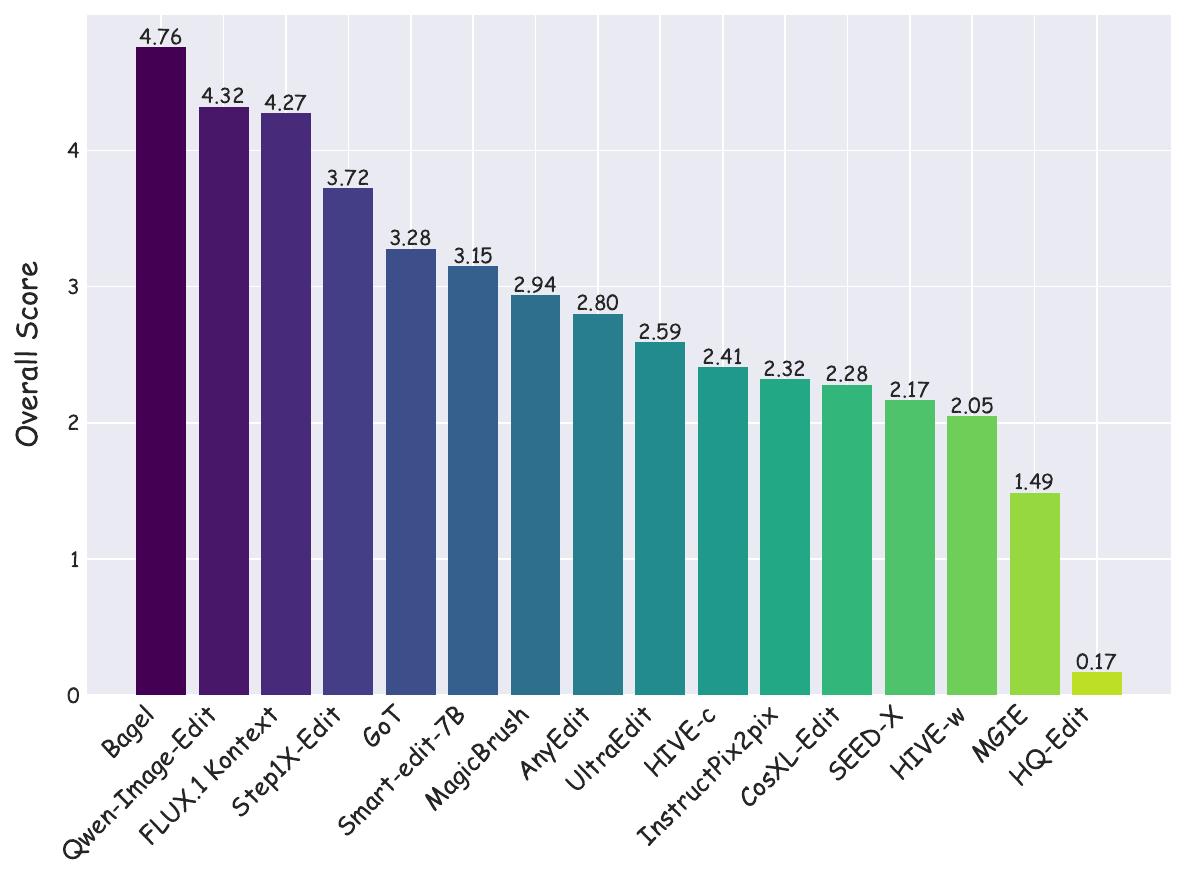}   % 原 figure1 文件名
    \caption{}                                    % 若需单独说明可写文字
    \label{fig:ranking}
  \end{subfigure}
  \hfill
  %-- 子图 (b) --

  \caption{\textbf{Overall Model Performance.}(a) Top 5 model performance in five major tasks. (b) Overall model performance across all tasks.}
  \label{fig:overall result}
\end{figure*}

\section{Insights}
In this section, we provide deeper insights and future directions based on model performances and failure analysis on our proposed CompBench.

\paragraph{The Critical Role of MLLMs.} We observe a strong correlation between architectural design and editing performance. As shown in Figure~\ref{fig:overall result}(a) and (b) (calculation details in supplementary), the top-performing models (Bagel~\citep{deng2025emerging}, Qwen-Image-Edit~\citep{wu2025qwen}, FLUX.1 Kontext~\citep{labs2025flux}, and Step1X-Edit~\citep{liu2025step1x}) mostly integrate multi-modal large language models (MLLMs). Models lacking MLLM integration frequently ignore instructions or edit wrong targets, indicating that standard CLIP-text alignment is insufficient for complex reasoning. Therefore, integrating MLLMs is indispensable for accurately interpreting intricate textual instructions and visual contexts.

\paragraph{Planner-Executor Misalignment.} Despite the advantages of MLLMs, our failure analysis reveals a widespread planner-executor misalignment. In high-density scenes, even when MLLMs (Planner) correctly identify targets, diffusion models (Executor) often fail at precise masking, causing background leakage. Unified architectures help mitigate this mismatch, as exemplified by Bagel's joint learning of understanding and generation. Consequently, future research should focus on improving pixel-level grounding stability in cluttered scenes, moving beyond purely semantic understanding.

\paragraph{Reasoning Bottleneck.} Beyond basic semantic alignment, complex multi-modal reasoning is foundational for high-fidelity edits. Enhancing reasoning capabilities significantly boosts performance, evident in data-centric strategies (e.g., SmartEdit training on LISA~\citep{lai2024lisa}'s reasoning segmentation data) and method-centric designs (e.g., GoT introducing Chain-of-Thought~\citep{wei2022chain} via MLLMs). Therefore, a crucial direction for future work involves continuously enhancing the reasoning capabilities of MLLMs through reasoning-aware training paradigms to ensure precise intent interpretation.

\paragraph{Geometric Hallucination.} Beyond semantics, we observe severe limitations in physical consistency. For complex spatial tasks like Action and Viewpoint editing, models frequently hallucinate distorted geometries. To address this issue, future frameworks need to incorporate 3D structural priors or geometric guidance into 2D-trained editors to maintain strict physical consistency during generation.
\section{Conclusion}
In this work, we introduce CompBench, the first large-scale benchmark designed for complex instruction-guided image editing. Our benchmark encompasses five major categories with nine specialized tasks comprising over 3,000 high-quality image editing pairs with corresponding instructions. We conduct extensive evaluation to systematically assess the capabilities and limitations of contemporary editing systems and validate our evaluation framework. Our findings not only reveal significant performance gaps in current models but also provide valuable insights to guide future research toward next-generation image editing systems.
\section*{Acknowledgements}
This work is supported by the National Natural Science Foundation of China (NO. 62572193), the Open Research Fund of the Key Laboratory of Advanced Theory and Application in Statistics and Data Science, Ministry of Education, and the Fundamental Research Funds for the Central Universities.
{
    \small
    \bibliographystyle{ieeenat_fullname}
    \bibliography{main}
}

% WARNING: do not forget to delete the supplementary pages from your submission 
% \input{sec/X_suppl}

\end{document}